\pdfoutput=1
\documentclass[11pt]{article}

\usepackage[a4paper,margin=1in]{geometry}
\usepackage{amsmath,amssymb,amsfonts}
\usepackage{graphicx}
\usepackage{booktabs}
\usepackage{multirow}
\usepackage{hyperref}
\usepackage[numbers]{natbib}
\usepackage{caption}
\usepackage{subcaption}
\usepackage{algorithm}
\usepackage{algpseudocode}
\usepackage{fancyhdr}
\usepackage{bm}

\hypersetup{
    colorlinks=true,
    citecolor=black,
    linkcolor=black,
    urlcolor=black
}

\title{\textbf{Multi-Level Analyzation of Imbalance to Resolve \\
Non-IID-Ness in Federated Learning}
\\[0.5em]
}

\author{
Haengbok Chung$^{1,2}$,
Jae Sung Lee$^{1,2,3}$\thanks{Corresponding Author. E-mail: jaes@snu.ac.kr}
}
\date{}

\begin{document}

\maketitle

\begin{center}
\footnotesize
$^{1}$Interdisciplinary Program in Artificial Intelligence,
Seoul National University, Republic of Korea\\
$^{2}$Department of Nuclear Medicine,
Seoul National University College of Medicine, Republic of Korea\\
$^{3}$Brightonix Imaging Inc. Seoul, Republic of Korea\\[0.3em]
\end{center}

\begin{center}
\footnotesize
Author Accepted Manuscript — Neurocomputing (2025)
\end{center}

\vspace{0.5em}

\noindent

\vspace{1em}

\begin{abstract}
\noindent Class imbalance is a common problem in deep learning that severely degrades performance. \textcolor{black}{In federated learning (FL), it is a critical factor contributing to non-identically distributed data (non-IID).} \textcolor{black}{Building on several previous attempts,} we define and analyze imbalance issues in FL at three levels: inter-case, inter-class, and inter-client. \textcolor{black}{Inter-case imbalance addresses the imbalance in every single class; inter-class imbalance compares the number of data between different classes. Inter-client imbalance represents different skewness of local data between clients.} \textcolor{black}{Based on these concepts, we propose FedBB, which consists of two main components:} (1) Positive Negative Balanced (PNB) loss function \textcolor{black}{addresses} the inter-case and inter-class imbalances in local \textcolor{black}{training, enhancing generalization on highly skewed local client datasets.} \textcolor{black}{It optimizes both multi-label and multi-class classifications by assigning higher weights to minority cases or classes.} (2) Client Balanced Reweighting (CBR) reweights clients based on inter-client imbalance during model aggregation, \textcolor{black}{giving greater weight to models trained on less skewed datasets.} Various experiments on X-ray and natural image datasets demonstrate that FedBB \textcolor{black}{outperforms} other algorithms in \textcolor{black}{both} performance and efficiency. \textcolor{black}{Additionally}, it requires limited statistical information, which is beneficial for privacy protection. Through ablation studies, we proved that PNB loss and CBR \textcolor{black}{independently contribute to performance}. \textcolor{black}{As FedBB aims to build a global model that accurately classifies all classes}, it can serve as a baseline for the generic and personalized FL.
\end{abstract}

\section{Introduction}

Federated learning (FL) \cite{1} is a promising collaborative learning method that preserves privacy \cite{2, 3, 4}. It allows many distributed clients to train their models in a distributed environment while protecting their data from being exposed outside the local nodes. In addition, because FL does not require large data centers, it can improve hardware efficiency and security compared to centralized training \cite{66}. 

However, the most critical problem of FL is that data distributions of clients are not independently or identically distributed (non-IID-ness) \cite{5}. As a result, the performance of FL deteriorates compared to conventional centralized training. Several algorithms have been proposed to solve this problem, including regularization methods \cite{6,7,8}, contrastive learning \cite{12}, knowledge distillation \cite{13,14,15}, personalization \cite{16,17,18}, meta-learning \cite{20,21}, and distribution fusion \cite{22}. However, these studies did not consider class imbalance, which is one of the significant reasons for non-IID-ness. Because class imbalance constitutes non-IID-ness, it becomes proportionally severe as the data distributions of clients become imbalanced. Besides, they required relatively large computation or resource overhead. Several studies \cite{23,24,25} recently have attempted to mitigate the negative impact of class imbalance. These methods are based on the conventional concept of class imbalance which directly compares the numbers of data between all categories. However, these works only focus on the multi-class classification. In addition, the concept of conventional class imbalance is insufficient for mitigating class imbalances in multi-label classification. \textcolor{black}{Specifically}, the features of two randomly sampled images \textcolor{black}{with} common categories can differ significantly if either one is included in an additional category. \textcolor{black}{For example, in chest X-rays, two images with similar conditions might appear vastly different if one has an additional condition, such as effusion or atelectasis, which significantly alters the lung's appearance.} Consequently, we must consider $2^n$ types of label sets \cite{65} in multi-label classification. From the Bayesian perspective, if $y_i$ is $i^{th}$ kind of label and $p(\cdot)$ denotes its probability, $p(y_i)$ \textcolor{black}{(the likelihood of $i^{th}$ category)} decreases exponentially owing to a higher number of categories. Therefore, labels with more categories remain relatively underrepresented, even if the conventional class imbalance is regulated.

Furthermore, the conventional method of aggregating clients' models is also problematic. It assigns larger weights to the models trained with larger datasets. However, because clients' datasets are highly skewed, this aggregation method results in a biased global model. This biased model \textcolor{black}{will likely} classify given input as the class that they have seen a lot \cite{28}. Because all clients have different data distributions, this bias deteriorates the overall fairness and reduces the generalization and personalization performance. 
	
To resolve these two problems, we divide class imbalance into three levels \textcolor{black}{to enhance both federated multi-label and multi-class classifications}: inter-case, inter-class, and inter-client. Inter-case imbalance considers the difference between the numbers of positive and negative cases, focusing on a single category. Examples not included in a particular category are considered negative cases for that category. When we calculate the binary cross-entropy loss, the number of negative cases is usually significantly greater than the positive cases. As a result, by mitigating the inter-case imbalance, we can prevent a model from being overfitted to the negative case of each class. The concept of inter-class imbalance \cite{29}, which compares the number of examples among the categories for a single client, is widely used for typical class imbalance problems. Inter-client imbalance analyzes the heterogeneous skewness among clients' datasets. The reason we analyze inter-client imbalance is to prevent the global model from overfitting to the majority classes of clients with large but skewed datasets.

Based on these concepts, we propose \textcolor{black}{\textbf{Federated Learning with Positive Negative Balanced loss and Client Balanced Reweighting (FedBB)}}, which optimizes both local learning and model aggregation. Our local learning goal is to create models that can accurately classify all classes. We enable this by introducing \textbf{Positive Negative Balanced (PNB)} loss function, which assigns higher weights to minor cases and classes by analyzing the inter-case and inter-class imbalances. The resulting local models can serve as \textcolor{black}{solid} baselines for generic and personalized FL \cite{30}. We optimize model aggregation by further analyzing the impact of class imbalance based on inter-client imbalance. We propose \textbf{Client Balanced Reweighting (CBR)}, which assigns higher weights to clients with less skewed datasets. This increases the generalizability of the global model by leveraging the limited statistical information, which is helpful for privacy protection.

\textcolor{black}{Various experiments were conducted on large-scale medical X-ray and natural image datasets to demonstrate the feasibility of the proposed FedBB}. 

The key contributions of this study are as follows:

\begin{itemize}
    \item \textcolor{black}{We propose Federated Learning with Positive Negative Balanced loss and Client Balanced Reweighting (FedBB), by analyzing} the multifaceted impacts of imbalance \textcolor{black}{issues in federated learning}.  We divide them into three levels: inter-case, inter-class, and inter-client. 
    \item We propose \textcolor{black}{Positive Negative Balanced (PNB)} loss function to mitigate both inter-case and inter-class imbalance problems in local learning. PNB loss assigns higher weights to the minority cases and classes.
    \item We introduce \textcolor{black}{Client Balanced Reweighting (CBR)}, which reweights local models based on the inter-client imbalance during model	aggregation. CBR evaluates local models trained with less skewed datasets are more valuable.
    \item \textcolor{black}{Through extensive experiments, we demonstrate that FeBB shows consistent superiority in terms of performance and efficiency in medical, natural image datasets with various experimental setups.}\\
\end{itemize}

\section{Related Work}
\textbf{FL Optimization.} Since the introduction of FL \cite{1}, several studies have aimed to improve its performance. \textcolor{black}{Regarding the variables in the following explanations, $\omega$ is the model, $t$ is the index of communication round, $i$ is the index of each client, and $\mu$ is the hyperparameter that regulates scale.}\\
\indent The first group of studies optimizes local training.  FedProx \cite{8} brings local models closer to the global model using the proximal term $ \frac{\mu}{2} ||\omega^{t} - \omega_i^{t}||^2$. Ditto \cite{10} extends FedProx by proposing a simple multitask learning framework that satisfies personalization, fairness, and robustness requirements. MOON \cite{11} employs contrastive learning to reduce the gap between the global model $\omega^t$ and the local model $\omega^t_i$ and to increase the gap between the previous round’s local model $\omega^{(t-1)}_i$ and the current one. They add the model-contrastive loss to the cross-entropy loss. FedAlign \cite{12, 31} regularizes local learning by leveraging the weight-sharing sub-block to prevent FL performance deterioration because of drifting clients. \textcolor{black}{ISFL \cite{77} adjusts the importance of each client's data dynamically through importance sampling, addressing the performance degradation that can occur due to non-IID data distributions.} \\
\indent The second group of FL studies focus on model aggregation \cite{18, 22, 32, 33, 78}. pFedLA \cite{18} performs layer-wise model aggregation by employing hypernetworks to improve personalization performance. \textcolor{black}{FedNova \cite{78} considers each client's number of local \textcolor{black}{updates and reflects} it when aggregating local models.} RHFL \cite{32} uses Symmetric Cross Entropy Learning (SL) loss to optimize FL performance in a noisy label environment. FedFusion \cite{22} infers global data distribution, which is composed of several components by leveraging a variational autoencoder (VAE).\\
\indent In addition to these approaches, several methods \cite{13, 14, 16, 79} have focused on optimizing the entire FL procedure. Distillation-based methods \cite{13, 14, 15} do not require parameter sharing; thus, they can protect against various attacks, enabling each client to use heterogeneous model architecture. RIPFL \cite{16} achieves reliability and interpretability for the personalized FL in multi-class classification for many clients. \textcolor{black}{FedOpt \cite{79} employs the Sparse Compression Algorithm (SCA) to reduce communication overhead while maintaining performance and enhancing security by integrating differential privacy and homomorphic encryption.} However, all these methods require relatively large resources or computation overhead. FedBB achieves state-of-the-art (SOTA) performance with small resource and computation overhead \textcolor{black}{as represented in Section \ref{effan}}. \textcolor{black}{FedBB uses hierarchical analysis to identify and address the imbalance issue by optimizing both the local training phase and the global model aggregation process}
\\
\\
\textbf{FL with Class Imbalance.} In the real world, data distributions are severely skewed in most cases. \\
\indent In FL, class imbalance causes non-IID-ness, which is the primary reason for performance deterioration. Several methods \cite{23, 24, 25, 26, 27, 69, 70, 71, 72, 73, 74, 75} recently have been introduced to resolve class imbalance in FL. FedIR \cite{24} is an importance reweighting scheme wherein the importance weights calculated according to the amount of global and local data are multiplied by the local objective function. Fed-ROD \cite{25} leverages balanced and empirical risk minimization loss functions for personalized and generic models. These models are merged in the model aggregation and then redistributed. CLIMB \cite{26} balances the impact of the majority and minority classes on the loss value, enforcing similar average loss values among clients. FedLC \cite{27} calibrates the logits by adding a pairwise label margin and applying the softmax function. The pairwise label margin is the smoothed difference between the numbers of data in the two classes. CReFF \cite{73} proposed classifier re-training on federated features, which results in a similar performance performance with the one re-retrained on real data. CLIP2FL \cite{75} leveraged CLIP for a novel FL framework \textcolor{black}{consisting of client-side} and server-side learning. \\
\indent This type of work is advantageous \textcolor{black}{because} it can be harmonized with many studies \cite{1, 8, 10, 11, 12}. However, these studies focus only on multi-class classification. In addition, some of them require relatively large computation or resource overhead. By contrast, we optimize both multi-label and multi-class classifications more efficiently. In addition, we analyze class imbalance at the inter-case, inter-class, and inter-client levels, whereas previous studies only consider inter-class imbalance. By leveraging the limited statistical information that restricts the server from accessing local data, our study protects \textcolor{black}{clients' privacy-sensitive raw data distributions from leaks.}

\noindent\textbf{Imbalanced Classification.} In real-world applications, severely skewed datasets are usually encountered. Therefore, several studies have attempted to address long-tailed (LT) data distributions \cite{52,53,54,55,56,57,58,59, 60}. In medicine, this imbalanced data distribution can have especially catastrophic consequences because it relates to lives. 
Consequently, addressing the problem due to imbalanced data distribution is \textcolor{black}{vital} to improve deep learning performance, particularly in the medical field. There have been some studies in which datasets were manipulated directly through re-sampling \cite{36} to balance data distribution. However, these methods have certain limitations. Over-sampling requires additional effort to train an additional generative model. Under-sampling risks losing important information included in the original dataset. Therefore, we leverage the re-weighting mechanism \cite{37, 38, 39} to further improve the generalizability of local training.

\section{Methodology}

\subsection{Problem Definition}

For both multi-label and multi-class classification, we set the cross-silo FL topology with a centralized server and $K$ distributed clients; thus, $|\mathcal{H}| = K$, where $\mathcal{H}$ denotes the client collection. Each client has a vector $\mathcal{A}_k$, which represents the amount of data in all classes for client $k$ with $|\mathcal{A}_k| = C$ \textcolor{black}{when there are C categories}, and $y_i^k$ denotes the one-hot vector of the ground truth with $|y_i^k| = C$. $i$ is the index of a sample. In addition, each client has a local model $\theta_k$ with weight $\mathcal{W}_k$ in the model aggregation. These weights are calculated before the local learning, and $f(\cdot)$ denotes the deep learning network. \textcolor{black}{The overall algorithm is summarized in \ref{Appendix:algorithm}.}

\subsection{Multi-Level Analyzation of Class Imbalance in FL}

Before explaining the PNB loss and CBR, we define the two levels of class imbalance as follows: 
\newline

\noindent\textbf{Definition 1 (Inter-case imbalance).} We define the probability that a random example belongs to the $i^{th}$ (1 $\leq$ $i$ $\leq$ C) category is $p(y_i)$ and $1 - p(y_i)$  for the opposite. The case that an example belongs to the $i^{th}$ category denotes the positive case of the $i^{th}$ category, whereas the latter denotes the negative case of the $i^{th}$ category. We define the discrepancy between these two cases as the inter-case imbalance. As a result, there are C number of inter-case imbalance values for each category. \textcolor{black}{We can represent this by the difference between $\alpha^k_{jp}$ and $\alpha^k_{jn}$ in the Equation \ref{eqn:6} when $k$, $j$ are the index of clients and categories and $p$, $n$ represent the positive and negative cases within a category.}\\

\noindent\textcolor{black}{\textbf{Definition 2 (Inter-class imbalance).} This imbalance is the difference between $p(y_i)$ values among categories, which has been regarded as a general class imbalance in previous studies \cite{29, 36, 37, 38, 39}. As a result, we have inter-class imbalance values corresponding to the number of clients. This also can be represented as $\alpha^k_{jp}$ when $j$ is the index of a category as in Equation \ref{eqn:6}. }\\

\noindent\textbf{Definition 3 (Inter-client imbalance).} The standard deviation of $p(y_i)$ (1 $\leq$ i $\leq$ C) can be calculated for each client. This value represents data skewness. We define the difference in these values between clients as an inter-client imbalance. Each client has multiple inter-case imbalance values corresponding to the number of classes but only one inter-client imbalance value. \textcolor{black}{It can be described by the difference of $\omega_k^a$ in the Equation \ref{eqn:11} when $k$ is the index of clients.} 
\\
\newline
Overall, inter-client imbalance is the most macroscopic imbalance compared to inter-case and inter-class imbalance. Inter-class imbalance is the middle-level imbalance between inter-case and inter-client imbalances. To mitigate non-IID-ness, we thought we should analyze the class imbalance from the micro to the macro scales rather than just focusing on the inter-class imbalance.

\subsection{Positive$-$negative$-$balanced (PNB) Loss}

We introduce a loss function that is designed to improve local learning performances which become the basic building block of FL. Thus far, several studies have focused on reweighting the inter-class imbalance. These studies reweighted the loss using inverse class frequency \cite{41} or its smoothed version \cite{42, 43}. Although it is common knowledge that inter-class imbalance deteriorates deep learning performance, to the best of our knowledge, inter-case imbalance has garnered little attention from researchers. \textcolor{black}{Therefore, we address inter-class and inter-case imbalances to further improve the training performance.}

\subsection{PNB Loss in Multi-label Classification }

In a multi-label classification task with $n$ categories, $2^{n}$ types of labels exist. However, few studies have focused on the imbalance of various label types. In the best-case scenario, similar amounts of data exist for each type to prevent model bias. However, from a Bayesian perspective, the probability of a specific label type decreases as the number of categories increases. Therefore, in addition to assigning higher weights to the minority categories, higher weights should \textcolor{black}{also be} assigned to samples that  \textcolor{black}{have more positive categories. For example, in chest X-rays, cases with multiple co-occurring diseases (categories), such as effusion and atelectasis, are rarer than those with a single or no disease, so we assign higher weights to such samples to improve classification performance.} To implement this insight, we introduce PNB loss, which reweights inter-case and inter-class imbalances. To begin with, we mitigate inter-case imbalance by analyzing the number of positive and negative cases for each class. 

Let us now review the binary cross-entropy (BCE) loss. Let's assume there is a single category. The numbers of positive and negative cases that contribute to the overall loss value can be observed in the following equation: 

\begin{equation}
\begin{split}
\label{eqn:1}
 \mathcal{L}_{BCE} = &  - [\sum_{j = 1}^{p}y_{j} * log(f(x_j)) \\
                    &  + \sum_{j=1}^{n}(1 - y_j) * log(1 - f(x_j))]
\end{split}
\end{equation}

Consequently, if we use BCE loss, the model is likely to be overfitted owing to the majority case. To prevent catastrophic overfitting, the overall impact of the positive and negative cases should be regulated more evenly. This can be implemented straightforwardly by multiplying the exact inverse frequencies of the positive and negative cases as follows:

\begin{equation}
\begin{split}
\label{eqn:2}
\mathcal{L}_{BCE'} = &\; - [\frac{1}{n}\sum_{j = 1}^{p}y_{j} * log(f(x_j)) \\
                    & + \frac{1}{p}\sum_{j=1}^{n}(1 - y_j) * log(1 - f(x_j))]
\end{split}
\end{equation}

However, multiplying by the exact inverse number of data as \cite{44} may degrade the performance. For instance, if certain categories of a dataset contain an extremely small amount of data, the impact of a single positive case on those categories will be significantly greater than that of a negative case, impairing the generalizability of the model. Therefore, the weighting numbers should be smoothed by using an effective number of samples, similar to \cite{43}. Expanding to multi-label classification with $C$ categories, the effective number of positive and negative cases in a category can be defined as follows:

\begin{equation}
\label{eqn:3}
E_{jp}^{k} = \frac{1 - \beta^{N_{jp}^{k}}}{1 - \beta},\; E_{jn}^{k} = \frac{1 - \beta^{N_{jn}^{k}}}{1 - \beta}
\end{equation}

$\beta$ denotes the hyperparameter, $N_{jp}^k = A_j^k/\tau$, $N_{jn}^k= B_j^k/\tau$, and $A_j^k$ and $B_j^k$ denote the actual numbers of positive and negative cases, respectively, in category $j$ for client $k$. Because $\beta$ would vanish if the amount of data is excessively large, we divide the actual amount of data with temperature $\tau$ to smoothen them. Thereafter, the equation is normalized as follows:

\begin{equation}
\label{eqn:4}
\begin{split}
\alpha_{jp}^{k} = \frac{(E_{jp}^{k})^{-1}}{(E_{jp}^{k})^{-1} + (E_{jn}^{k})^{-1}}, \\
\alpha_{jn}^{k} = \frac{(E_{jn}^{k})^{-1}}{(E_{jp}^{k})^{-1} + (E_{jn}^{k})^{-1}}
\end{split}
\end{equation}

Using these terms, we express the smoothed inter-case reweighting loss function as follows: 

\begin{equation}
\label{eqn:5}
\begin{split}
\mathcal{L} = &-[\sum_{j = 1}^{C}(\alpha_{jp}^{k}\sum_{i = 1}^{p_j}y_i^k * log(f(x_i^k)) + \\
              &\alpha_{jn}^{k}\sum_{i=1}^{n_j}(1 - y_i^k) * log(1 - f(x_i^k)))]
\end{split}
\end{equation}

$C$ denotes the number of categories.
Finally, by multiplying $\alpha_{jp}^k$ again with the overall loss, we can mitigate inter-class imbalance can be regulated. Because $\alpha_{jp}^k$ is from the number of positive cases in each class, it is proportional to the number of data in each class. Therefore through multiplying this value to $(\alpha_{jp}^{k}\sum_{i = 1}^{p_j}y_i^k * log(f(x_i^k)) + \alpha_{jn}^{k}\sum_{i=1}^{n_j}(1 - y_i^k) * log(1 - f(x_i^k)))$, we can regulate the contribution of loss values from different classes. As a result, we resolved inter-case and inter-class imbalances. $\mu$ is the scaling hyperparameter; the final equation of the PNB loss is as follows:

\begin{equation}
\label{eqn:6}
\begin{split}
\mathcal{L}_{PNB} = & - [\sum_{i=1}^{N^k}\sum_{j = 1}^{C}(\mu * \alpha_{jp}^{k} * \\
                        & (\alpha_{jp}^{k} * y_i^k * log(f(x_i^k)) + \\
                        & (\alpha_{jn}^{k} * (1 - y_i^k) * log(1 - f(x_i^k))))]
\end{split}
\end{equation}

Consequently, the PNB loss assigns higher weights that are inversely proportional to the label probability. 

\begin{figure}[h]
\begin{center}
\includegraphics[width=2in]{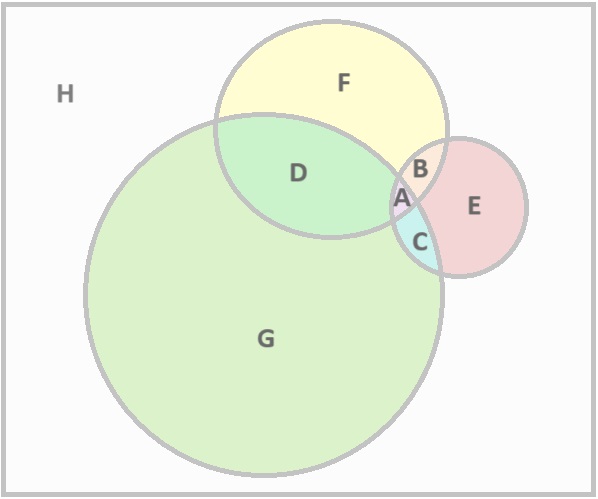}
\caption{Illustration of various types of labels of multi-label classification with three categories.}
\label{fig:pnb}
\end{center}
\end{figure}

To further elucidate PNB loss, we introduce a simple example. Assume that there are three different categories. In Figure \ref{fig:pnb}, the various types of labels are denoted as $A–H$ and we set $p(A) < p(B) < p(C) < p(D) < p(E) < p(F) < p(G) < p(H)$. 

The PNB loss assigns \textcolor{black}{higher} weights to the loss values of samples from minority cases or categories. For the samples from the majority case or classes, PNB loss assigns relatively lower weights. We define the former as upweighting and the latter as downweighting. For the convenience of explanation, let's assume the negative cases for all categories are always the overwhelming majority case so the differences between their assigned weights are meaningless. Then, we can set $\lambda$ as a common inter-case upweighting value for all positive cases. Thus, $\nu$ can serve as a common downweighting value for all negative cases. $\lambda$ and $\nu$ is related to the $\alpha_{jp}^{k}$ and \textcolor{black}{$\alpha_{jn}^{k}$} in the $(\alpha_{jp}^{k}\sum_{i = 1}^{p_j}y_i^k * log(f(x_i^k)) + \alpha_{jn}^{k}\sum_{i=1}^{n_j}(1 - y_i^k) * log(1 - f(x_i^k)))$ \textcolor{black}{because they regulate inter-case imbalances}. The larger value among $\alpha_{jp}^{k}$ and \textcolor{black}{$\alpha_{jn}^{k}$} becomes $\lambda$ and the smaller one becomes $\nu$. \textcolor{black}{This is because when the larger and smaller values compared to the median values among $\alpha_{jp}^{k}$ and $\alpha_{jn}^{k}$ equal to the relative upweighting downweighting terms.} In addition, we can set the inter-class imbalance upweighting parameter to $\rho$, and the downweighting parameter to $\psi$. The $\alpha_{jp}^{k}$ which is multiplied to the $(\alpha_{jp}^{k}\sum_{i = 1}^{p_j}y_i^k * log(f(x_i^k)) + \alpha_{jn}^{k}\sum_{i=1}^{n_j}(1 - y_i^k) * log(1 - f(x_i^k)))$ is related to the $\rho$ and $\psi$. \textcolor{black}{In this case, when the $\alpha_{jn}^{k}$ is grater than the median $\alpha_{jn}^{k}$, this equal to the upweighting $\rho$ and when $\alpha_{jn}^{k}$ is smaller compared to the median value among them, this relates to the downweighting $\psi$. In other words, when} $j$ indicates the majority class (the number of data in the class is greater than the average number of data of all classes), it can be regarded as $\rho$ and for $\psi$, vice versa.

When $l(\cdot)$ denotes the overall loss value of a given label type, the reweighting result is obtained as follows: 

\begin{equation}
\label{eqn:7}
\begin{split}
\mathcal{L}_{PNB} = &\lambda^3\rho\psi * l(A) + \lambda^2\rho\nu * l(B) + \lambda^2\rho\psi\nu * l(C) \\
                    & + \lambda^2\psi\nu * l(D) + \lambda\rho\nu^2 * l(E) + \lambda\nu^2 * l(F)\\
                    & + \lambda\psi\nu^2 * l(G) + \nu^3 * l(H)
\end{split}
\end{equation}

Naturally, because $\lambda > \rho > \psi > \nu$, the magnitudes of the coefficients become inversely proportional to the probability of label types. Thus, the loss values from minority cases or categories repeatedly upweighted and for the opposite, vice versa. Although the real cases can be more complicated, PNB loss eventually balances the impact of various types of labels to the overall loss value based on this fundamental. The experimental results of PNB loss for multi-label classification using medical chest X-ray data are represented in Table \ref{tab:7}.

\begin{figure}
\begin{center}
\includegraphics[width=5.5in]{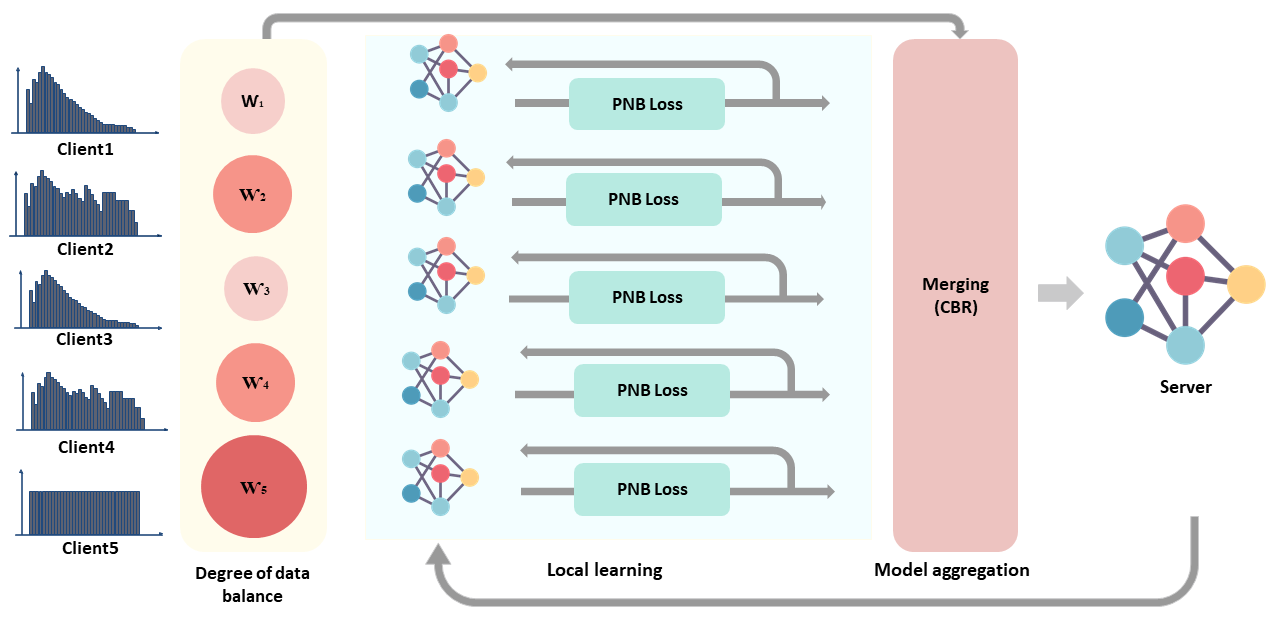}
\end{center}
\caption{FedBB process comprises CBR and PNB loss functions. The weights used in the model aggregation are pre-calculated based on each client’s data skewness. PNB loss and CBR aims to make a better generalized global model.}
\label{fig:FedBB}
\end{figure}

\subsection{PNB Loss in Multi-class Classification}

Because multi-class classification is based on the cross-entropy loss, we can express the PNB loss in Equation (\ref{eqn:6}) as follows:  

\begin{equation}
\label{eqn:8}
\mathcal{L}_{PNB} = \sum_{i=1}^{N^k}\sum_{j = 1}^{C} \mu * \alpha_{jp}^{k} *  y_i^k * log(x_i^k)
\end{equation}

\begin{table}
\begin{center}
\begin{tabular}{cc}
\hline
\multicolumn{2}{c}{\textbf{CIFAR-10}} \\
\hline
Imbalance Factor & Accuracy \\
\hline
0.8 & 82.91 \\
0.5 & 81.42\\
0.2 & 76.64\\
\hline
\end{tabular}
\end{center}
\caption{Different test accuracies on the CIFAR-10 dataset depending on the imbalance factor. Imbalance factor represents the proportion of the amount of data between the maximum and minimum classes.}
\label{tab:1}
\end{table}

where the variables are obtained through inter-case imbalance reweighting and $\mu$ is a hyperparameter used to regulate the scale. 

It should be noted that the PNB loss in multi-class classification resembles the CB loss \cite{43}, a method developed to address inter-class imbalance. However, the reweighting terms in the PNB loss are obtained from the inter-case imbalance calculation. Both $E_{jp}^k$ and $E_{jn}^k$ are required to calculate $\alpha_{jp}^k$. In addition, unlike the CB loss, we scale the loss value through the hyperparameter $\mu$, whose value is generally between 1 and 5 based on the experimental conditions. Through the experiments described in Experimental Results, we found that setting an appropriate $\mu$ value is more crucial for improving FL performance, particularly in multi-class classification tasks than in multi-label ones. In fact, the PNB loss is significantly more powerful than the CB loss as it boosts the training performance by considering broader and more detailed aspects of the reweighting mechanism. In addition, CB loss resulted in relatively poor centralized training performance than the PNB loss, as demonstrated by the results presented in Table \ref{tab:7}.

\begin{table*}[!ht]
\begin{center}
\resizebox{\textwidth}{!}{
\begin{tabular}{c|ccc|c|ccc|c}
\hline
 & \multicolumn{4}{|c|}{\textbf{NIH CXR14}} & \multicolumn{4}{c}{\textbf{CheXpert}} \\
\hline
 & \multicolumn{3}{|c|}{\textcolor{black}{Performance $\uparrow$}} & \textcolor{black}{Efficiency $\downarrow$} & \multicolumn{3}{|c|}{\textcolor{black}{Performance $\uparrow$}} & \textcolor{black}{Efficiency $\downarrow$} \\
\hline
Algorithm & $\delta = 1$ & $\delta = 0.5$ & $\delta = 0.1$ & Time & $\delta = 1$ & $\delta = 0.5$ & $\delta = 0.1$ & Time\\
\hline
FedAvg & 69.02 $(\pm0.20)$ & 69.02 $(\pm0.13)$ & 67.77 $(\pm0.04)$ & 1311m & 67.75 $(\pm0.20)$ & 66.59 $(\pm0.49)$ & 65.60 $(\pm0.24)$ & 913m\\
FedProx & 68.40 $(\pm0.06)$ & 68.33 $(\pm0.50)$ & 68.51 $(\pm0.30)$ & 1392m & 68.06 $(\pm0.16)$ & 66.36 $(\pm0.30)$ & 65.42 $(\pm0.55)$& 1023m\\
MOON & 66.30 $(\pm0.30)$ & 64.08 $(\pm0.14)$ & 66.67 $(\pm0.28)$ & 33100m & 66.17 $(\pm0.16)$ & 65.07 $(\pm0.16)$ & 63.29 $(\pm0.14)$ & 33450m\\
FedNova & 66.92 $(\pm0.0.58)$ & 66.82 $(\pm0.34)$ & 66.64 $(\pm0.20)$ & 1311m & 68.23 $(\pm0.11)$ & 67.70 $(\pm0.31)$ & 65.08 $(\pm2.30)$ & 915m\\
FedAlign & 67.57 $(\pm0.35)$ & 68.46 $(\pm0.07)$ & 67.61 $(\pm0.23)$ & 1365m & 66.25 $(\pm0.10)$ & 65.50 $(\pm0.16)$ & 64.89 $(\pm0.01)$ & 974m\\
\hline
\textbf{FedBB (ours)} & \textbf{71.78 $\mathbf{\bm{(\pm0.06)}}$} & \textbf{70.50 $\bm{(\pm0.35)}$} & \textbf{72.47 $\bm{(\pm0.43)}$} & 1317m & \textbf{68.90$\bm{(\pm0.07)}$} & \textbf{68.39 $\bm{(\pm0.04)}$} & \textbf{68.84 $\bm{(\pm0.20)}$} & 931m\\
\hline
\end{tabular}}
\end{center}
\caption{\textcolor{black}{AUCs and time measurements} for the NIH CXR14 and CheXpert datasets. Each dataset was split using $K = 5$ and different $\delta$ values ($\delta = 1$, $\delta = 0.5$ and \textcolor{black}{$\delta = 0.1$}). We set 50 model aggregations and 2 local epochs, the value of the $\tau$ was set to 10, and $\beta$ was set to 0.9999999. \textcolor{black}{We measured the time it took to conduct the experiment in the $\delta = 0.1$ setting. We measured time using NVIDIA 4070RTX GPU. $m$ represents minute.}}
\label{tab:50}
\end{table*}

\begin{table}[!ht]
\begin{center}
\begin{tabular}{c|cccc|cccc}
\hline
 & \multicolumn{4}{|c|}{\textbf{NIH CXR14}} & \multicolumn{4}{c}{\textbf{CheXpert}} \\
 \hline
 & \multicolumn{2}{c}{E = 2} & \multicolumn{2}{c|}{E = 5} & \multicolumn{2}{c}{E = 2} & \multicolumn{2}{c}{E = 5} \\
\hline
Algorithm & $\delta = 1$ & $\delta = 0.5$ & $\delta = 1$ & $\delta = 0.5$ & $\delta = 1$ & $\delta = 0.5$ & $\delta = 1$ & $\delta = 0.5$ \\
\hline
FedAvg & 67.54 & 67.65 & 68.10 & 64.80 & 67.34 & 67.97 & 65.40 & 64.31 \\
FedProx & 66.57 & 66.97 & 67.34 & 65.74 & 67.12 & 67.82 & 65.58 & 65.18\\
MOON & 64.53 & 65.73 & 67.02 & 64.22 & 67.40 & 66.65 & 63.64 & 61.28\\
FedNova & 65.97 & 65.32 & 67.60 & 67.33 & 66.28 & 66.25& 66.03 & 66.12\\
FedAlign & 65.55 & 67.20 & 65.77 & 66.89& 62.40 & 66.82 & 64.35 & 63.42\\
\hline
\textbf{FedBB (ours)} & \textbf{ 71.14 } & \textbf{ 71.29 } & \textbf{72.02} & \textbf{68.98} & \textbf{ 69.31 } & \textbf{ 69.34 } & \textbf{66.20}& \textbf{69.37}\\
\hline
\end{tabular}
\end{center}
\caption{\textcolor{black}{AUCs} for the NIH CXR14 and CheXpert datasets. We set 30 model aggregations and 2, \textcolor{black}{5} local learning epochs (E). The value of the $\tau$ was set to 1, \textcolor{black}{and} $\beta$ was set to 0.9999. The other conditions are \textcolor{black}{the} same as Table \ref{tab:50}.}
\label{tab:2}
\end{table}

\begin{table}
\begin{center}
\begin{tabular}{ccc}
\hline
 & \textbf{NIH CXR14} & \textbf{CheXpert} \\
\hline
CZ & 78.39 & 82.40\\
\hline
\end{tabular}
\end{center}
\caption{Best AUC result of centralized training during 20 epochs.}
\label{tab:CZ}
\end{table}

\begin{table*}
\begin{center}
\begin{tabular}{c|ccc|ccc}
\hline
\multicolumn{5}{c}{\textbf{CIFAR10}}\\
\hline
& \multicolumn{3}{c|}{\textcolor{black}{Performance $\uparrow$}} & \textcolor{black}{Efficiency $\downarrow$}\\
\hline
Algorithm & $\delta = 0.05$ & $\delta = 0.1$ & $\delta = 0.5$ & Time\\
\hline
FedAvg & 64.19 $(\pm0.26)$ & 76.81 $(\pm0.10)$ & 86.14 $(\pm0.46)$ & 165m\\
FedProx & 66.22 $(\pm0.33)$& 77.06 $(\pm0.56)$ & 85.92 $(\pm0.32)$ & 264m\\
MOON & 65.12 $(\pm0.19)$ & 76.88 $(\pm0.29)$ & 86.91 $(\pm0.15)$ & 430m\\
FedNova & 59.59 $(\pm0.06)$ & 74.90 $(\pm0.28)$ & 86.86 $(\pm0.08)$ & \textbf{97m} \\
FedAlign & 61.89 $(\pm1.07)$ & 75.39 $(\pm0.85)$ & 84.98 $(\pm0.23)$ & 430m\\
FedLC & \textbf{68.61 $(\bm{\pm0.29})$}& 77.35 $(\pm0.17)$ & 87.66 $(\pm0.05)$ & 264m\\
\hline
\textbf{FedBB (ours)} & 68.23 ($\pm0.95$)& \textbf{78.00 ($\bm{\pm0.56}$)} & \textbf{88.78 $\bm{(\pm0.26)}$} & 188m\\
\hline
\end{tabular}
\end{center}
\caption{Accuracies \textcolor{black}{and time measurements} for the CIFAR-10. $K = 10$ and the number of epochs for local training and model aggregation were 10 and 20. \textcolor{black}{We tested performance with different skewness ($\delta = $ 0.5, 0.1, and 0.05). We measured time using NVIDIA 4070RTX GPU. $m$ represents minute.} Smaller values indicate higher \textcolor{black}{efficiency in the Time column}.}
\label{tab:3_1}
\end{table*}

\begin{table*}
\begin{center}
\begin{tabular}{c|cc|cc}
\hline
 & \multicolumn{2}{c|}{\textbf{CIFAR100}} & \multicolumn{2}{c}{\textbf{Tiny-ImageNet}}\\
\hline
Algorithm & $\delta = 0.1$ & $\delta = 0.5$ & $\delta = 0.1$ & $\delta = 0.5$ \\
\hline
FedAvg & 53.33 $(\pm0.08)$ & 59.26 $(\pm0.13)$ & 36.86 $(\pm0.84)$ & 41.99 $(\pm0.17)$\\
FedProx & 53.29 $(\pm0.24)$ & 59.89 $(\pm0.24)$ & 36.12 $(\pm0.26)$& 41.14 $(\pm0.72)$\\
MOON & 53.11 $(\pm0.27)$ & 60.92 $(\pm0.08)$ & 35.42 $(\pm0.08)$& 41.97 $(\pm0.05)$\\
FedNova & 54.04 $(\pm0.49)$ & 60.49 $(\pm0.41)$ & 36.32 $(\pm0.10)$ & 42.02 $(\pm0.19)$\\
FedAlign & 54.43 $(\pm0.06)$ & 60.70 $(\pm0.47)$ & 32.74 $(\pm0.01)$& 19.08 $(\pm18.58)$\\
FedLC & 53.54 $(\pm0.15)$ & 60.50 $(\pm0.25)$ & - & -\\
\hline
\textbf{FedBB (ours)} & \textbf{57.24 $\bm{(\pm0.22)}$} & \textbf{63.14 $\bm{(\pm0.07)}$}& \textbf{37.70 $\bm{(\pm0.08)}$}&\textbf{43.01 $\bm{(\pm0.05)}$}\\
\hline
\end{tabular}
\end{center}
\caption{Accuracies for the CIFAR-100, Tiny-ImageNet datasets. $K = 16$ and the number of epochs for local training and model aggregation were 20 and 30.}
\label{tab:3_2}
\end{table*}

\begin{table}
\begin{center}
\begin{tabular}{c|cc|cc}
\hline
 & \multicolumn{2}{c|}{\textbf{LT-CIFAR10}} & \multicolumn{2}{c}{\textbf{LT-CIFAR100}}\\
\hline
Algorithm & $\delta = 0.1$ & $\delta = 0.5$ & $\delta = 0.1$ & $\delta = 0.5$ \\
\hline
FedAvg & 57.81 & 76.32 & 35.52 & 39.75 \\
FedProx & 62.02 & 74.16 & 35.01 & 40.18 \\
MOON & 58.14 & 77.16 & 36.78 & 41.02 \\
FedNova & 57.54 & 75.67 & 36.72 & 38.98\\
FedAlign & 59.92 & 72.99 & 39.27 & 41.37 \\
FedLC & \textbf{63.00} & \textbf{78.34} & \textbf{35.51} & \textbf{41.08} \\
\hline
\textbf{FedBB (ours)} & \textbf{64.30} &  \textbf{78.17}  & \textbf{ 40.05 } & \textbf{ 45.13 }\\
\hline
\end{tabular}
\end{center}
\caption{Accuracy on artificially made LT-CIFAR-10 and LT-CIFAR-100. Training setups are same as the original CIFAR-10/100.}
\label{tab:4}
\end{table}

\subsection{Client Balanced Reweighting (CBR)} 

We introduce a weighting scheme \textcolor{black}{that} is used for aggregating local clients' \textcolor{black}{models}. Our methodology can be regarded as an application of \textcolor{black}{the} aggregation mechanism of FedAvg. The difference is that our method \textcolor{black}{calculates} the weights which are multiplied by the model parameters based on the inter-class imbalance rather than \textcolor{black}{the} data amount.

The background insight of our method is as follows. \cite{45} explains that merging models can produce unwanted results owing to the loss barrier. In addition, the typical method of aggregating models involves reweighting according to the number of datasets, which can produce results that are biased toward the majority categories. 

Therefore, we aim to develop a more reliable model aggregation method. As stated in previous studies \cite{41, 42, 43}, a balanced data distribution typically results in more generalized models than the long-tailed data distribution. The results presented in Table \ref{tab:1} also demonstrate that the test accuracy is higher for \textcolor{black}{the} more balanced dataset. Based on this, we believe that to create a global model that performs well for every class, we should rely on local models with better generalization performances. 

From this perspective, we propose CBR, which assigns higher weights to clients with less skewed data distributions in model aggregation. \textcolor{black}{A client's data skewness is estimated by calculating the standard deviation of the amount of data in each category,} which can be regarded as the inter-class imbalance. This value varies among clients, producing the inter-client imbalance described previously. 

Using vector $\mathcal{D}_k$ ($|\mathcal{D}_k| = C$), we calculate the degree of data balance. $\mathcal{A}_k$ denotes the vector containing the number of data in all classes for client $k$. Thereafter, we subtract the mean of $\mathcal{A}_k(\mu_\mathcal{A})$ from each element in vector $\mathcal{A}_k$. As we intend to estimate the absolute distance using the mean and make it more noticeable, we square each element as follows: 

\begin{equation}
\label{eqn:9}
\mathcal{D}_k = (\mathcal{A}_k- \mu_\mathcal{A})^2        
\end{equation}

Subsequently, we \textcolor{black}{normalized} the distance vector and calculated the sum, which is a scalar value representing the degree of data imbalance. By considering the reverse of $\mathcal{R}_k$, we can express the degree of data balance as follows:

\begin{equation}
\label{eqn:10}
\mathcal{R}_k = \sum_{i = 1}^{C}(\frac{\mathcal{D}_{ki}}{\sum_{j = 1}^{C}\mathcal{D}_{kj}})   
\end{equation}

Because $\mathcal{R}_k$ is between 0 to 1, $\mathcal{R}_k^{-1}$ exceeds 1 which leads to gradient explosion when we repeatedly aggregate the local models in the central server. Therefore, we again normalize $\mathcal{R}_k$ to maintain its scale to be between 0 to 1 as follows:

\begin{equation}
\label{eqn:11}
\omega_{k}^{a} = \frac{\mathcal{R}_{k}^{-1}}{\sum_{i=1}^{K}\mathcal{R}_{i}^{-1}}
\end{equation}

This term can be used to confirm the $k^{th}$ client’s weight term. $\omega_k^b$ is defined as the weight based on the amount of data for client $k$, same as FedAvg. Using $\gamma$, the impacts of $\omega_k^a$ and $\omega_k^b$ can be controlled. The reason we use both $\omega_k^a$ and $\omega_k^b$ is to handle extreme cases. For instance, if a client has a balanced but scarce data distribution, using only $\omega_k^a$ would degrade performance. However, if a client’s dataset is large but highly skewed, assigning a higher weight to that client would degrade the FL performance. Therefore, the $k^{th}$ client’s weight during the model aggregation can be computed as follows:

\begin{equation}
\label{eqn:12}
\mathcal{W}_k = \gamma * \omega_k^a + (1 - \gamma) * \omega_{k}^{b}
\end{equation}

The effectiveness of CBR is demonstrated in Table \ref{tab:8}. The overall procedure of FedBB is illustrated in Figure \ref{fig:FedBB}.

\subsection{Convergence Proof of FedBB}

\cite{61} proved the convergence of FedAvg based on several assumptions. Through applying their work, we provide the convergence result bound of FedBB. In Eq. \ref{eqn:con}, the upper value represents the bound of FedBB, and the below term is the bound of FedBB. \textcolor{black}{Compared} to FedAvg, because FedBB has \textcolor{black}{a} smaller variance of stochastic gradients in each device, the bound of the convergence result of FedBB is smaller than other algorithms \cite{1, 6, 8, 11, 12}. Detailed explanation in Appendix \ref{Proof}.

\begin{equation}
\label{eqn:con}
\begin{split}
E[F(\omega_T)] - F^* & \leq \frac{\mathcal{K}}{\gamma + T + 1}(\frac{2B'}{\mu} + \frac{\mu\gamma}{2}E||\omega_1 - \omega^*||^2)\\
                     & \leq \frac{\mathcal{K}}{\gamma + T + 1}(\frac{2B}{\mu} + \frac{\mu\gamma}{2}E||\omega_1 - \omega^*||^2)
\end{split}
\end{equation}

\section{Experimental Results}

\subsection{Experimental Setup}
\textbf{Datasets.} We conducted experiments using X-ray image (NIH chestX-ray14, CheXpert \cite{48}) and natural image (CIFAR-10/100 \cite{51}, Tiny-ImageNet \cite{64}) datasets. We artificially made the quantity skew based on random sampling for the multi-label classification datasets (X-ray datasets) and created a label distribution skew for the multi-class classification datasets (CIFAR-10/100) \cite{49}. \\
\textbf{Implementation Details.} The images were distributed using a Dirichlet distribution which is widely used in previous studies. $\delta$ was set to control the degree of non-IID-ness, and a higher $\delta$ value implies lower non-IID-ness. For multi-label classification, because there are $2^n$ label types when there are $n$ classes, Dirichlet distribution is used to skew the quantity of data without considering label type. We calculate each client's proportion of data amount then distribute data according to that proportion. For example, when \textcolor{black}{there} is a total of 1000 samples, and three clients' proportions are 0.1, 0.3, and 0.6, each client \textcolor{black}{has} 100, 300, and 600 samples regardless of label types. On the other hand, for multi-class classification, we applied Dirichlet distribution for each class respectively. Therefore, each class \textcolor{black}{has a different} data distribution. As a result, while all clients may have similar data distribution in the multi-label environment, in multi-class classification, each client has a distinctive data distribution. Regarding the hyperparameter, the results were obtained with optimal $\mu$ values of 0.001, 1.0, and 0.45 for FedProx, MOON, and FedAlign, respectively \cite{8, 11, 12}. \textcolor{black}{These} values are from their original papers. For FedLC, ranging from 10 to 15, we assigned $\tau$ values which generate \textcolor{black}{the} best performance. For all methods, we employed ResNet56 \cite{50} and a Stochastic gradient descent (SGD) optimizer with a momentum value of 0.9. The learning rate was set to 0.01 for all experiments.  \textcolor{black}{The} batch size was set to 32 for FL and 64 for centralized training. For MOON, we added two projection layers to the ResNet56, as described in the study. \textcolor{black}{For the data pre-processing, we normalized data.} We experimented with two different random seeds. Regarding the hardware, we used NVIDIA 3090 GTX with 64GB RAM, 12th Gen Intel(R) Core(TM) i7-12700F. \textcolor{black}{We leveraged two random seeds (0, 1996) to obtain all results.}\\
\textbf{Evaluation Metrics.} For the X-ray image datasets, we leveraged the area under curve (AUC) of the Receiver Operating Characteristic Curve (ROC curve) to evaluate classification performance because it is an important and widely used metric in the medical domain. For the natural image datasets, we evaluated classification accuracy.

\subsection{Chest X-ray Images Classification}
Although FL has various applications, it is \textcolor{black}{imperative} in the medical field because of its inherent privacy concerns. \textcolor{black}{Therefore, testing the algorithm's efficacy on medical datasets is important. Occasionally, an algorithm may exhibit extremely different performances for medical and natural image datasets. Therefore, verifying the method's efficacy on medical datasets is necessary.}

To test the generalizability of FedBB for chest X-ray image datasets, we used two datasets from different domains: NIH chestX-ray14 (NIH CXR14) and CheXpert. These are long-tailed chest X-ray images. NIH CXR14 consists of only frontal views, while CheXpert consists of frontal and lateral views. These datasets contain some ambiguous labels, which were excluded (“No Finding,”  “Support Devices,” “Pleural Effusion,” and “Pleural Other”). Because the image size of NIH CXR14 is too large, we resized them to 150 × 150 pixels. We distributed 86,336 images to five clients for training and used 22,366 for the test.
\textcolor{black}{For FedBB}, $\mu$ \textcolor{black}{in PNB loss} was set to 4.0, and $\gamma$ was set to 1. \textcolor{black}{Regarding the $\alpha$, $\beta$ values, we set it to 10 and 0.9999999 for all experiments.}

As summarized in Table \ref{tab:50},\ref{tab:2}, FedBB achieved the SOTA AUC in all experiments. Although most algorithms could not outperform FedAvg for X-ray datasets, FedBB performed successfully for X-ray datasets under various conditions. \textcolor{black}{The superior performance of FedBB can be attributed to its comprehensive approach in addressing a wide range of imbalance issues in federated learning (FL), while other algorithms tend to focus on specific aspects—such as inter-class imbalance (FedLC) or inter-client imbalance (FedProx, FedNova, MOON, FedAlign). Although FedBB incurs additional computational overhead compared to FedNova or FedAvg, due to the calculation of imbalance factors, coefficients, and the extra multiplications for PNB loss, the increased complexity remains within a manageable range.}

\subsection{Natural Images Classification}

In addition to medical datasets, we compared the performances of FedBB on CIFAR-10/100, \textcolor{black}{and} Tiny-ImageNet with that of other algorithms because it is also important to deal with privacy concerns in domains that use natural images. CIFAR-10/100 consists of 50,000 training images and 10,000 test images. We distributed training samples to 10 and 16 clients, set local epochs to 10, \textcolor{black}{and} 20, and model aggregation epochs to 20 and 30 for CIFAR-10/100, following the \cite{12}. Tiny-ImageNet has 64 x 64 size 100,000 training images. We used 10,000 validation images for the test. We distributed these images to 16 clients and set local epochs and model aggregation epochs as 20 and 30.

In addition, we tested our algorithm on artificially created LT-CIFAR-10/100 datasets because this is common in real cases where the amount of data for some classes is scarce. This allowed us to \textcolor{black}{examine the proposed method's performance on a highly skewed dataset.} We set the imbalance factor to 0.1, implying that the number of data in the minimum class is ten times smaller than that in the maximum class. 

Regarding the hyperparameters \textcolor{black}{for FedBB}, we set $\mu$ in the PNB loss to 5.0 in the experiment using LT, original CIFAR-100, \textcolor{black}{CIFAR-10 dataset with $\delta = 0.05$} and 4.0 in the remainder of the experiments. Regarding the $\beta$ value, we set 0.9999 for CIFAR-10, 0.99 for CIFAR-100, 0.999 for LT-CIFAR100 \textcolor{black}{0.6 for CIFAR-10 with $\delta = 0.05$}. \textcolor{black}{About $\alpha$ value, we set it to 10 for LT-CIFAT100 ($\delta = 0.5$) and 1 for the rest of the experiments.} In addition, $\tau$ and $\gamma$ were set to 1 for CIFAR-10/100.

\textcolor{black}{Table \ref{tab:3_1}, \ref{tab:3_2}, \ref{tab:4}} summarizes the experimental results of FedBB and other algorithms, wherein FedBB achieved the highest accuracy with good efficiency. \textcolor{black}{The reason for FedBB's superior performance is the same as in the X-ray data experiments.}

Regarding the FedLC, we omitted its results from the Tiny-ImageNet because it failed to function with very low accuracies of less than 10. We assume the reason behind this result is FedLC is suitable for the relatively small number of data because it achieved meaningful performance on the CIFAR10, LT-CIFAR10, CIFAR100, and LT-CIFAR100. They use a constant (-1/4) in the calibration procedure, but it limits its adaptability in various conditions.

\begin{table*}
\begin{center}
\begin{tabular}{c|cccc|ccc}
\hline
\multicolumn{8}{c}{\textbf{FedBB}}\\
\hline
Dataset & $\alpha = 0.1$ & $\alpha = 1$ & $\alpha = 10$ & $\alpha = 100$& $\beta = 0.99$ & $\beta = 0.999$ & $\beta = 0.9999$\\
\hline
CIFAR10 & 88.78 & 88.72 & 88.26 & 88.40 & 87.93 & 88.54 & 88.26\\ 
LT-CIFAR10 & 77.62 & 78.17 & 76.97 & 77.93 & 76.99 & 74.59 &78.17\\ 
CIFAR100 & 62.75 & 62.78 & 63.00 & 63.29 & 63.00 & 63.02 &62.31\\ 
LT-CIFAR100 & 43.52 & 43.12 & 44.56 & 44.37 & 44.56 & 45.13 & 44.92\\ 
\hline
\end{tabular}
\end{center}
\caption{\textcolor{black}{We tested performance with different $\alpha$ ($\alpha = $ 0.1, 1, and 10). In these cases, the $\beta$, $\mu$, and $\delta$ were set to 0.9999, 4.0, and 0.5. We also tested performance with different $\beta$ ($\beta = $ 0.999, 0.9999, and 0.99999). We set $\alpha$, to 10 for CIFAR-100, LT-CIFAR100 and 1 for CIFAR-10, LT-CIFAR10. Regarding $\mu$, $\delta$, we set them to 4.0, and 0.5 in these experiments.}}
\label{tab:10}
\end{table*}

\begin{table*}
\begin{center}
\begin{tabular}{c|cc|c}
\hline
\multicolumn{4}{c}{\textbf{LT-CIFAR100}}\\
\hline
Algorithm & SOLO & w FedLC & \textbf{w FedBB (ours)}\\
\hline
FedAvg  & 39.66 & 41.08 & \textbf{43.05}\\
FedProx & 39.78 & 40.13 & \textbf{44.02}\\
MOON & 41.02 & 39.61 & \textbf{44.63}\\
FedAlign & 42.37 & - & \textbf{44.89}\\
\hline
\end{tabular}
\end{center}
\caption{Results with and without harmonization of FedLC, and FedBB. SOLO indicates the case that the algorithm is used without harmonization.}
\label{tab:6}
\end{table*}

\begin{table*}
\begin{center}
\begin{tabular}{c|cc}
\hline
\textbf{Algorithm} & $\mathbf{\#}$ \textbf{of Parameters} & \textbf{MFLOPs}\\
\hline
FedAvg & \textbf{0.61} & \textbf{87.3}\\
FedProx & 1.21 & \textbf{87.3}\\
MOON & 2.21 & 262.2\\
FedNova & \textbf{0.61} & 95.24 \\
FedAlign & \textbf{0.61} & 89.1\\
FedLC& \textbf{0.61} & \textbf{87.3}\\
\hline
\textbf{FedBB (ours)}  & \textbf{0.61} & \textbf{87.3}\\
\hline
\end{tabular}
\end{center}
\caption{\textcolor{black}{We compared the numbers of parameters and MFLOPs}. \textcolor{black}{The 'Param' column shows the number of network parameters required for each algorithms, and 'M' represents million. MFLOPs refers to the number of floating-point operations executed per second, where 1 MFLOP equals 1 million floating-point operations.}}
\label{tab:efficiency}
\end{table*}

\subsection{Ablation Study}

\textcolor{black}{We tested FedBB with various $\alpha$ and $\beta$ values to analyze their impact. For CIFAR-10 and LT-CIFAR10, we set $K = 10$, with 10 epochs for local training and 20 epochs for model aggregation. The imbalance factor for LT-CIFAR10 and LT-CIFAR100 was 0.1. For CIFAR-100 and LT-CIFAR100, we set $K = 16$, with 20 epochs for local training and 30 epochs for model aggregation. Table \ref{tab:10} presents the results of these experiments. As shown, while different combinations of $\alpha$ and $\beta$ affected performance, there were no clear trends or rules regarding these parameters. Therefore, to optimize FedBB's performance, experimenting with various $\alpha$ and $\beta$ values is necessary.}

Because FedBB optimizes the loss function and model aggregation, it can be used simultaneously with other algorithms. As summarized in Table \ref{tab:6}, FedBB successfully harmonized with different algorithms, resulting in higher test accuracies, while FedLC failed to harmonize with FedAlign. The experimental setting is \textcolor{black}{the same} as the Table \ref{tab:6}. The difference is that the FedAvg, FedProx, MOON, and FedAlign are used with the FedBB.

\textcolor{black}{We also verified (1) the independent effectiveness of PNB loss and CBR and (2) the effectiveness of FedBB in another experimental setup with a larger client number and low participant rate in \ref{appendix:further}. The overall performance plots are provides in the \ref{Appendix:plots}.}

\subsection{\textcolor{black}{Efficiency Analysis}}
\label{effan}
\textcolor{black}{We also analyze resource and computation efficiency by analyzing the number of parameters and MFLOPs required for all datasets. As summarized in Table \ref{tab:efficiency}, FedBB requires the same amount of memory and computation as FedAvg to achieve the best performance, which are 0.6 million of parameters and 87.3 million of floating point of operations. It is three-times more smaller compared to the MOON which requires 2.21 million of parameters and 262.2  million of floating point operations.}

\section{Conclusion}
The non-IID-ness attributed to class imbalance deteriorates FL performance and several studies have attempted to resolve this problem by addressing the class imbalance problem. In this study, we extended these trials by proposing FedBB, which divides the problem of class imbalance into three levels: inter-case, inter-class, and inter-client. For local learning, we introduced the PNB loss function, which regulates inter-case and inter-class imbalances. For model aggregation, we proposed CBR, which assigns higher weights to local models trained with less skewed datasets based on the inter-client imbalance.\\
\indent Various experiments on X-ray and natural image datasets demonstrate FedBB performs better than other algorithms with generalizability and efficiency. \textcolor{black}{Carefully setting hyperparameters in FedBB was required to achieve these performances.} In addition, it requires limited statistical information, which is beneficial for privacy protection. \textcolor{black}{We discussed the further ways to prevent privacy leakage in the \ref{Appendix:privacy}.} In addition, the PNB loss can be leveraged for both FL and centralized training. Because the objective of FedBB is to create a global model that can accurately classify every class, it can serve as a baseline for the generic and personalized FL. \\
\indent However, FedBB requires a centralized server. In addition, because it is a parameter-based approach, it is vulnerable to attacks. Therefore, future studies should investigate a serverless, nonparameter-based method to overcome these limitations.

\section{Acknowledgement}
This work was supported by K-Brain Project (No. RS-2023-00264160) and the Basic Research Project (NO. RS-2024-00354123) of the National Research Foundation (NRF) funded by Ministry of Science and ICT, the Korea Medical Device Development Fund grant funded by the Korea government (the Ministry of Science and ICT, the Ministry of Trade, Industry and Energy, the Ministry of Health \& Welfare, the Ministry of Food and Drug Safety) (Project Number: 1711137868, RS-2020-KD000006), and the Institute of Information \& communications Technology Planning \& Evaluation (IITP) grant [NO.RS-2021-II211343, Artificial Intelligence Graduate School Program (Seoul National University)]. 

\section{\textcolor{black}{Data availability statement}}
\noindent \textcolor{black}{NIH ChestXray14: https://www.kaggle.com/datasets/nih-chest-xrays/data}\\
\textcolor{black}{CheXpert: https://www.kaggle.com/datasets/ashery/chexpert}\\
\textcolor{black}{CIFAR-10, CIFAR-100: https://cs.toronto.edu/~kriz/cifar.html}\\
\textcolor{black}{Tiny-ImageNet: https://www.kaggle.com/c/tiny-imagenet/data}


%%       \citet{<label>} ==> Jones et al. [21]
%%       \citep{<label>} ==> [21]
%%

% \bibliographystyle{elsarticle-num-names} 
\bibliographystyle{unsrt}
\bibliography{references}

\newpage

\appendix
\section*{Appendix}
\addcontentsline{toc}{section}{Appendix}

\section{Foreword}
This section includes several detailed explanations of our study. First, we illustrate the impact of regulating inter-case and inter-class imbalance. Second, verifies the effectiveness of CBR through additional experiments. Through these experiments, we can see that CBR successfully improved the performance. Third, we conduct experiments to analyze the FedBB performance with a relatively large number of clients and low participant rates. Fourthly, we search for ways to prevent privacy leakage for a more secure implementation of FedBB. Fifthly, we show FedBB's overall algorithm. Sixthly, we prove the convergence result bound of FedBB. Seventhly, we argue our contribution to the healthcare community which requires a specific environment to execute FL. Eighthly, we show the plots of the experiments on various datasets.

\section{Analysis of PNB Loss}
\begin{figure*}
\begin{center}
\includegraphics[width=1.8in]{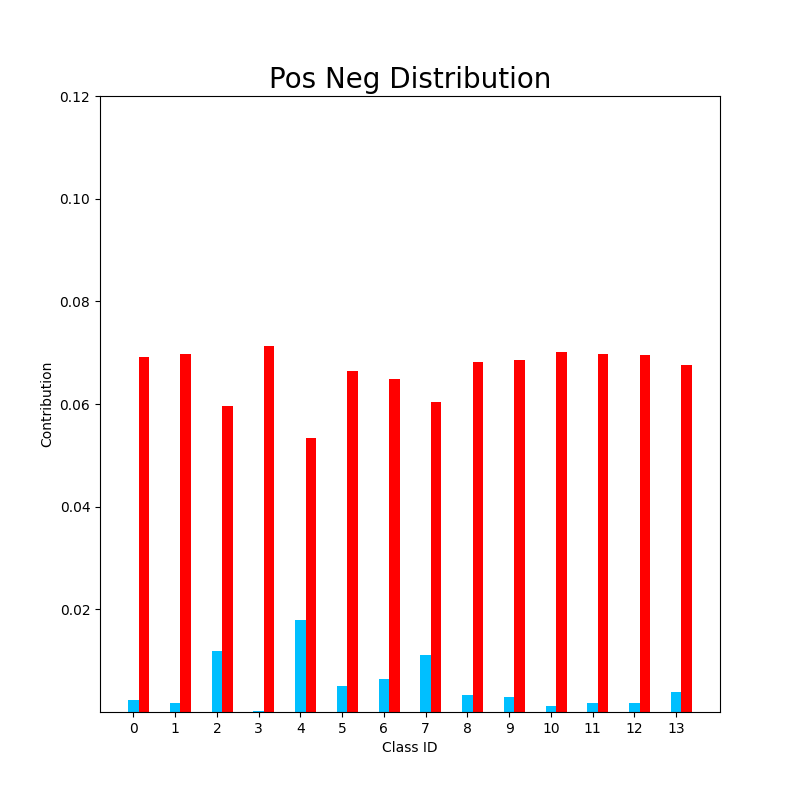}
\includegraphics[width=1.8in]{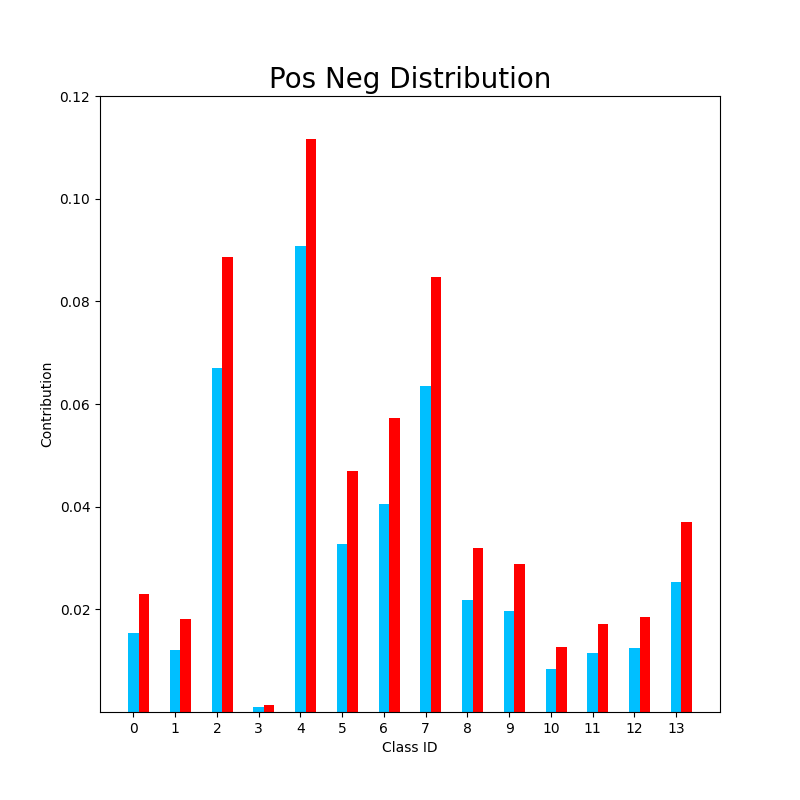}
\includegraphics[width=1.8in]{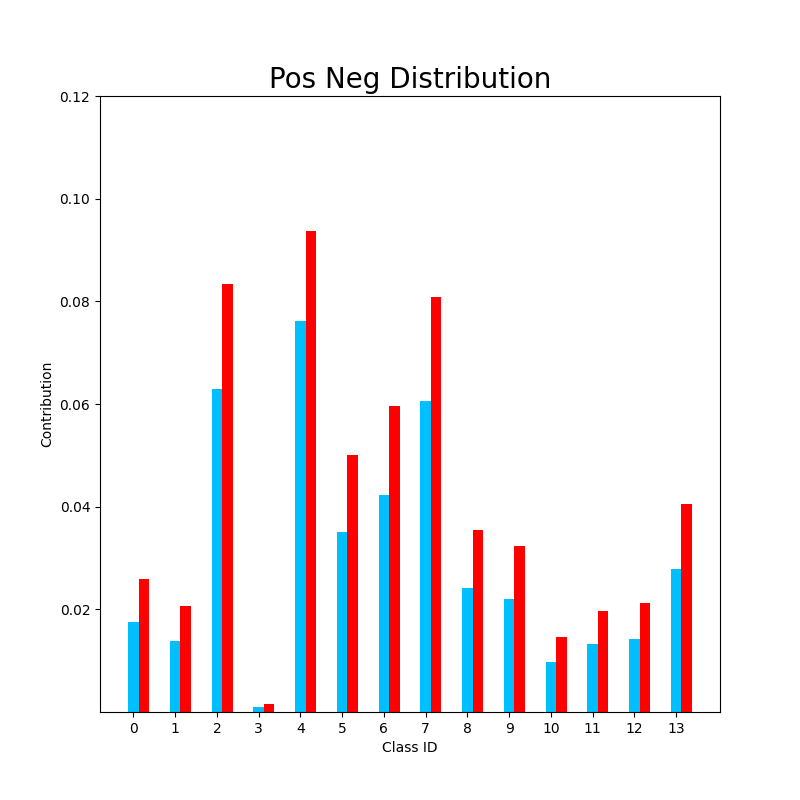}
\end{center}
\caption{Blue bars denote the positive rate of the overall dataset and red bars denote the negative rate of all classes. All classes’ sum of the positive and negative contributions is 1. The Left shows each class’s original contribution to the overall loss value. After balancing inter-case imbalance (middle). Finally, after balancing both inter-case and inter-class imbalances, the contribution to the loss value becomes more balanced (right).}
\label{fig:dis}
\end{figure*}

We visualize the effect of regulating inter-case and inter-class imbalances. Figure.\ref{fig:dis} shows the changes of each case's and class's impact on the overall loss value before and after regulating. The used dataset is NIH CXR14.

\section{\textcolor{black}{Further Ablation Study}} 
\label{appendix:further}
\subsection{\textcolor{black}{\textbf{Verification of Independent Effectiveness of PNB loss and CBR}}}
We verify the effectiveness of PNB loss and CBR, respectively. 

\textcolor{black}{For the PNB loss, we tested the performance of PNB loss for NIH CXR14\textcolor{black}{, LT-CIFAR10} and LT-CIFAR100 to verify its efficacy. \textcolor{black}{We set the imbalance factor as 0.01 and 0.1 in LT-CIFAR10 and LT-CIFAR100. Regarding the $\beta$, we set it as 0.9999999, 0.999, 0.9999 respectively. $\alpha$ was 10, 1, 1 in three cases.} Although we tested FedLC with various $\tau$ values ranging from 0.0001 to 1000, it failed to function in the centralized training. In contrast, as \textcolor{black}{summarized} in Table \ref{tab:7}, PNB loss achieved \textcolor{black}{the} highest performance in centralized training.}

\begin{figure*}
\begin{center}    
\includegraphics[width=1.8in]{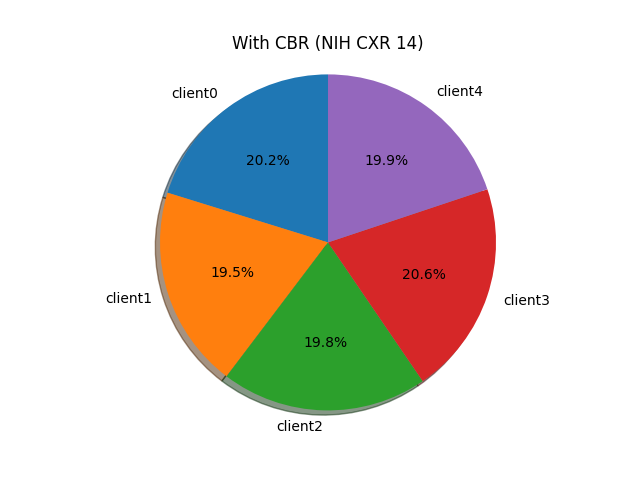}
\includegraphics[width=1.8in]{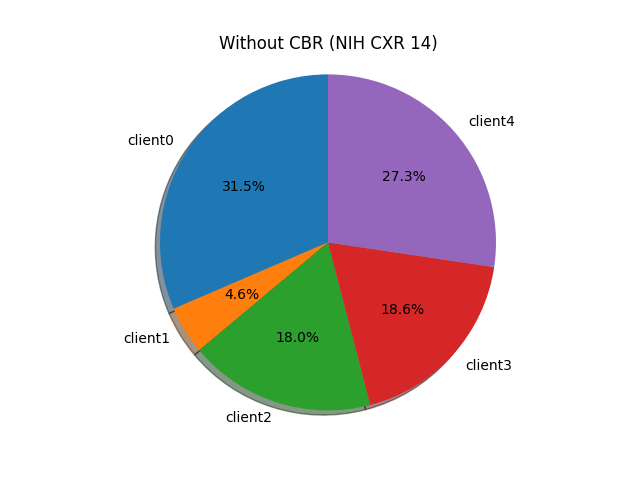}
\includegraphics[width=1.8in]{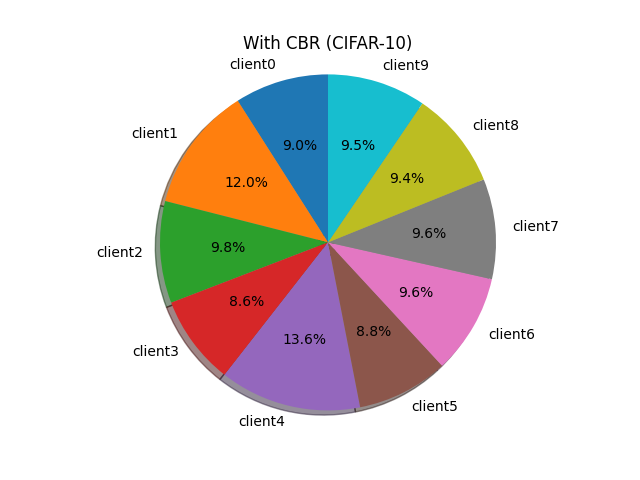}
\includegraphics[width=1.8in]{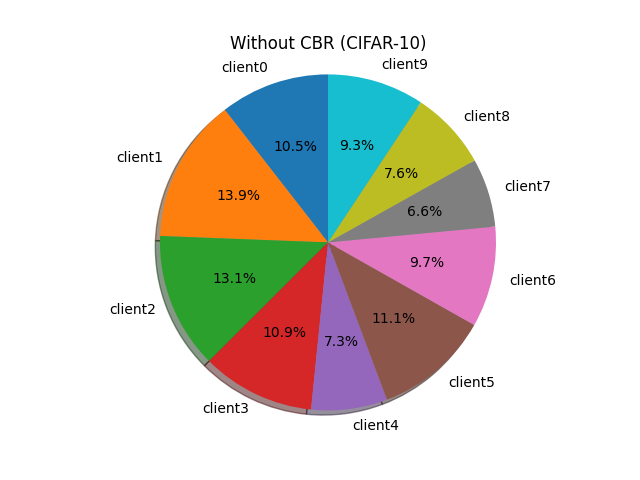}
\includegraphics[width=1.8in]{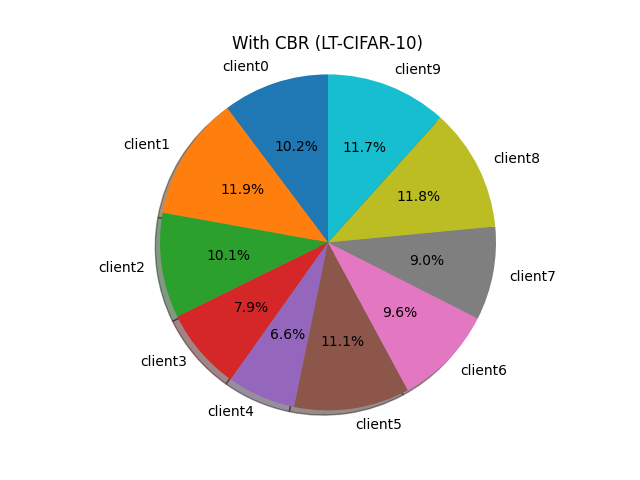}
\includegraphics[width=1.8in]{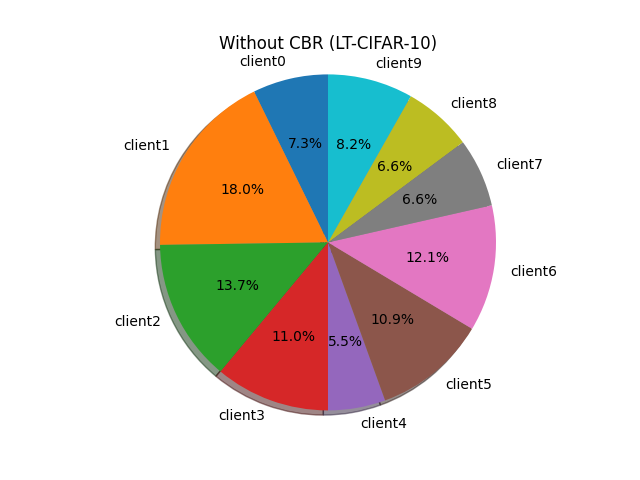}
\end{center}
\caption{Proportions of clients' models in the model aggregation stage. NIH CXR 14, CIFAR-10, and LT-CIFAR-10 are used to compare these values. The values are changed after applying CBR.}
\label{fig:CBR}
\end{figure*}

For the experiments to prove the efficiency of CBR, the number of local training epochs for the medical X-ray datasets is two, and the model aggregation epoch is 50. Regarding the experiments using natural image datasets, the number of local training epochs is 10 \textcolor{black}{and} 20, and the number of model aggregation epochs is 20 \textcolor{black}{and} 30 for CIFAR-10 and CIFAR-100, respectively. \textcolor{black}{We can see that CBR successfully improved the performances through the experiments in various conditions.} Figure.\ref{fig:CBR}
 shows the proportions of clients' models in the model aggregation round. After applying CBR, the proportions are changed according to their data skewness.

\begin{table}[!ht]
\begin{center}
\begin{tabular}{c|c|cc}
\hline
 & \textbf{NIH CXR14} & \textbf{LT-CIFAR10} & \textbf{LT-CIFAR100}\\
\hline
Loss & AUC & \multicolumn{2}{|c}{ACC}\\
\hline
BCE Loss & 76.86 & - & - \\
CE Loss & - & 48.46 & 40.23 \\
Focal Loss & 76.71 & 47.89 & 38.72\\
CB Loss & 75.34 & 49.78 & 40.71\\
FedLC Loss & - & 10.00 & 4.66\\
\hline
\textbf{PNB Loss (ours)} & \textbf{78.39} & \textbf{55.12} & \textbf{42.72}\\
\hline
\end{tabular}
\end{center}
\caption{Best AUC over 20 epochs for NIH CXR14 and final test accuracy over 30 epochs for \textcolor{black}{LT-CIFAR10 and }LT-CIFAR-100 in centralized training. In this experiment, EfficientNetB0 was used as the model for NIH CXR14 and ResNet56 for LT-CIFAR-100.}
\label{tab:7}
\end{table}
 
\begin{table}[!ht]
\begin{center}
\begin{tabular}{c|cc|cc}
\hline
\textbf{Dataset} & \multicolumn{2}{c}{\textbf{w/o CBR}} & \multicolumn{2}{|c}{\textbf{w CBR}} \\
\hline
 & $\delta = 1$ & $\delta = 0.5$ & $\delta = 1$ & $\delta = 0.5$ \\
\hline
NIH CXR 14 & 71.12 & 69.33 & \textbf{71.84} & \textbf{70.85} \\
CheXpert & 68.46 & 66.32 & \textbf{68.83} & \textbf{68.43}\\
\hline
 & $\delta = 0.5$ & $\delta = 0.1$ & $\delta = 0.5$ & $\delta = 0.1$\\
 \hline
CIFAR-10 & 87.91 & 77.81 & \textbf{89.04} & \textbf{78.56}\\
LT-CIFAR-10 & 76.47 & 62.99 & \textbf{78.17} & \textbf{64.30}\\
CIFAR-100 & 62.62 & 56.98 & \textbf{63.21} & \textbf{57.46}\\
LT-CIFAR-100 & 42.83 & 38.62 &\textbf{43.13} & \textbf{40.05}\\
\hline
\end{tabular}
\end{center}
\caption{Results of FedBB with and without applying CBR. \textcolor{black}{We leveraged PNB loss in all cases.}}
\label{tab:8}
\end{table}

\begin{figure*}[t]
\begin{center}
\includegraphics[width=1.8in]{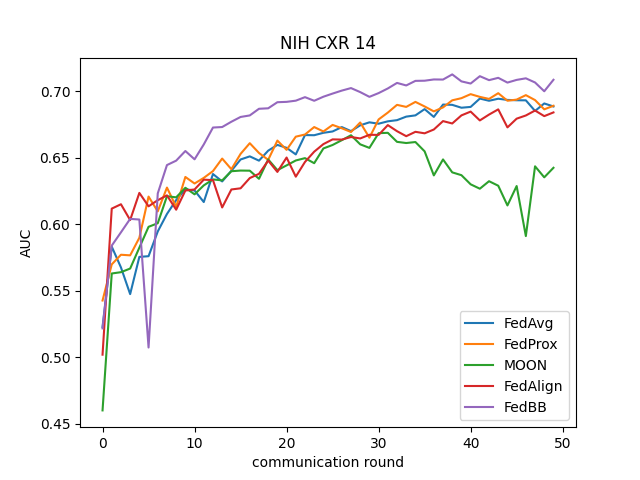}  \includegraphics[width=1.8in]{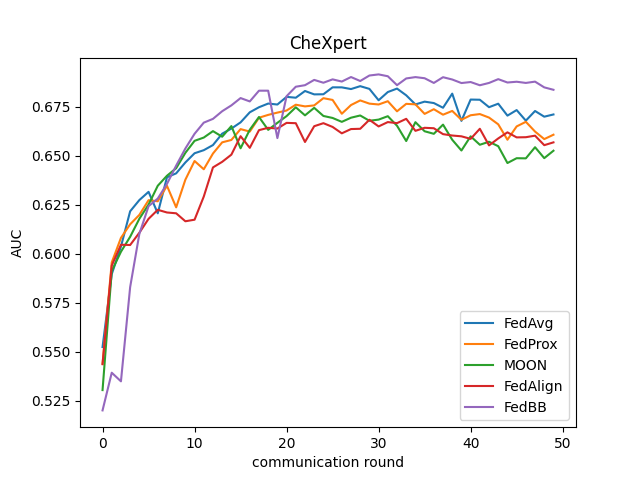}  \includegraphics[width=1.8in]{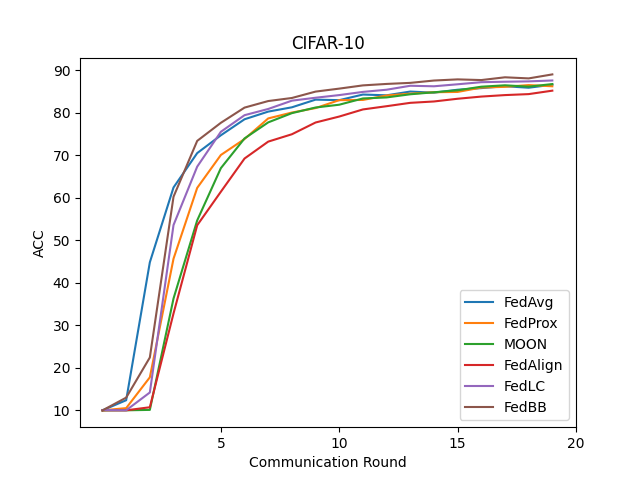}\\
\includegraphics[width=1.8in]{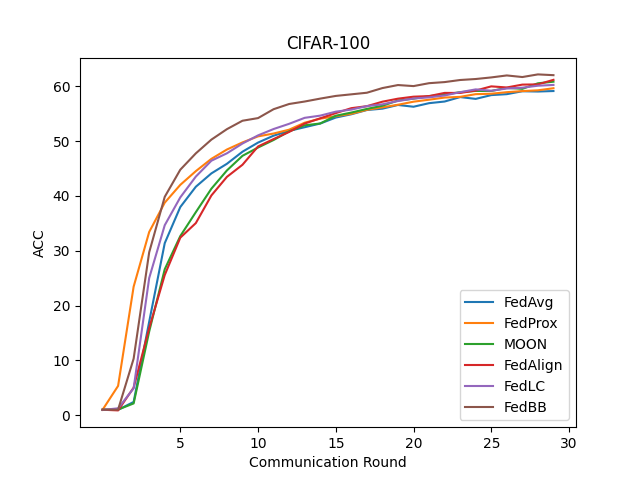}  \includegraphics[width=1.8in]{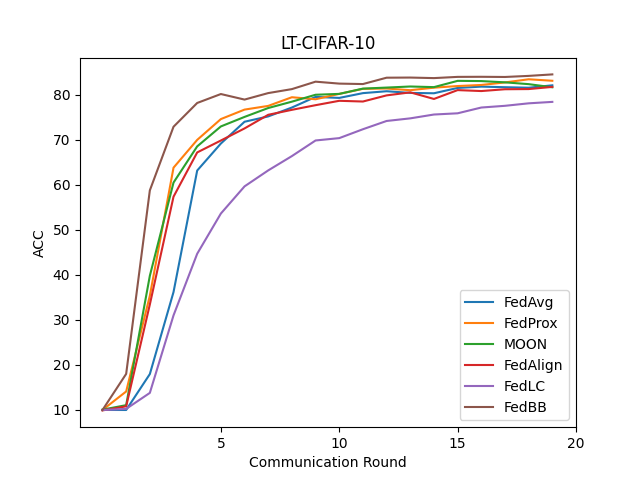} \includegraphics[width=1.8in]{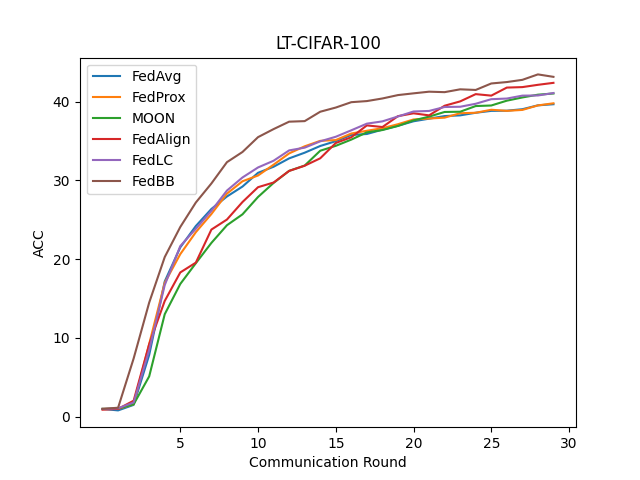}
\end{center}
\caption{Test AUC plots of NIH CXR14 and CheXpert and the test accuracy plots of CIFAR-10/100 and LT-CIFAR10/100. FedBB outperformed other algorithms in a stable manner.}
\label{fig:result}
\end{figure*}

\subsection{\textcolor{black}{\textbf{Performance Verification with Large Client Number and Low Participant Rate}}}
We aim to prove the effectiveness of FedBB in another environment \textcolor{black}{with a large client number and low participant rate.} $\mu$ was set to 5.0 in FedBB. In the experiments using medical X-ray data, the number of local epochs and model aggregation rounds were set to 2 and 50. In CIFAR-10, they were set to 10 and 50. For CIFAR-100, they were set to 20 and 30. \textcolor{black}{As shown in Table \ref{tab:9}, FedBB proved its superiority in terms of AUC and accuracy compared to other algorithms.}

\begin{table*}[!ht]
\begin{center}
\begin{tabular}{c|cc|cc}
\hline
 & \textbf{NIH CXR14} & \textbf{CheXpert} & \textbf{CIFAR-10} & \textbf{CIFAR-100}\\
\hline
Algorithm & \multicolumn{2}{c|}{AUC} & \multicolumn{2}{c}{ACC}\\
\hline
FedAvg & 57.14 & 61.31 & 69.77 & 33.38 \\
FedProx & 57.46 & 62.05 & 70.88 & 32.54\\
MOON & 57.28 & 61.69 & 71.83 & 28.63 \\
FedNova & 57.25 & 60.45 & 65.35 & 31.11\\
FedAlign & 61.94 & 60.66 & 71.48 & 32.74\\
FedLC & - & - & 73.56 & 32.67\\
\hline
\textbf{FedBB (ours)} & \textbf{64.87} & \textbf{63.91} & \textbf{73.64} & \textbf{36.72}\\
\hline
\end{tabular}
\end{center}
\caption{\textcolor{black}{The performances with a relatively large number of clients and low participant rate. We set the number of client and participant rates to 100 and 0.05, respectively.} $\delta$ is 1 for NIH CXR 14, CheXpert and 0.5 for CIFAR-10/100.}
\label{tab:9}
\end{table*}

\section{Ways to Prevent Privacy Leakage}
\label{Appendix:privacy}
Although CBR does not expose raw information about clients' data distribution, there could be a concern that this partial information leakage is also dangerous. Therefore, we suggest ways to exchange this partial information confidentially. First, we can apply cryptographic techniques such as secure multi-party computation (MPC) with homomorphic encryption (HE). \textcolor{black}{We can prevent hackers from intervening in information transactions by executing certification procedures.} In addition, we can set up a trusted execution environment (TEE). This enables only qualified participants \textcolor{black}{to} see the raw information about the client's data distribution. On the other hand, differential privacy (DP), which is a widely used security mechanism, is not a good solution for CBR because it directly changes the value, which can lead to sub-optimal model aggregation. Although the \textcolor{black}{methods above} cause some computation overheads, we can improve safety by preventing sensitive data distribution information from being directly exposed \textcolor{black}{to} outside clients.

\section{Performance Plots}
\label{Appendix:plots}
We illustrate the overall performance of FedBB by comparing other algorithms. In Figure \ref{fig:result}, we can see that FedBB converged faster than other algorithms in a stable manner. 

\section{Algorithm}
\label{Appendix:algorithm}
Algorithm \ref{alg:algorithm1} shows how to calculate weights that are multiplied by the clients' model parameters. This procedure is done before the training starts. The calculation of PNB loss is represented in the Algorithm \ref{alg:algorithm2}. The weights multiplied by each term in the PNB loss \textcolor{black}{are} pre-calculated before the training starts as CBR. The overall algorithm of FedBB is explained in Algorithm \ref{alg:algorithm3}.

\begin{algorithm}[htb]
\caption{Weight calculation for CBR}
\label{alg:algorithm1}
\textbf{Input}: $\mathcal{A}_k$ is the vector containing the number of data in all classes for client $k$, $\mu_\mathcal{A}$ is the mean of $\mathcal{A}_k$, $k$ is the index of clients, $C$ is the number of classes, $K$ is the number of clients, $\gamma$ is hyperparameter.
\begin{algorithmic}[1] %[1] enables line numbers

\State $\omega_{k}^{b} = \frac{sum(\mathcal{A_k})}{\sum_{i=1}^K sum(\mathcal{A_i})}$ $\#$ same as FedAvg

\State $\mathcal{D}_k = (\mathcal{A}_k- \mu_\mathcal{A})^2$

\State $\mathcal{R}_k = \sum_{i = 1}^{C}(\frac{\mathcal{D}_{ki}}{\sum_{j = 1}^{C}\mathcal{D}_{kj}})$ $\#$ degree of data imbalance

\State $\omega_{k}^{a} = \frac{\mathcal{R}_{k}^{-1}}{\sum_{i=1}^{K}\mathcal{R}_{i}^{-1}}$ $\#$ set the scale of $\omega_{k}^{a}$ to be between 0 to 1.

\State $\mathcal{W}_k = \gamma * \omega_k^a + (1 - \gamma) * \omega_{k}^{b}$

\end{algorithmic}
\end{algorithm}

\begin{algorithm}[htb]
\caption{PNB loss function}
\label{alg:algorithm2}
\textbf{Input}: $\beta$, $\mu$, $\tau$ are the hyperparameters. $p$ is the positive case and $n$ is the negative case. $k$ is the index of clients, $j$ is the index of categories. $N_{jp}^k = A_j^k/\tau$ , $N_{jn}^k= B_j^k/\tau$, and $A_j^k$ and $B_j^k$ denote the actual numbers of positive and negative cases.

\begin{algorithmic}[1]
\State $E_{jp}^{k} = \frac{1 - \beta^{N_{jp}^{k}}}{1 - \beta},\; E_{jn}^{k} = \frac{1 - \beta^{N_{jn}^{k}}}{1 - \beta}$ $\#$ effective number calculation
\State $\alpha_{jp}^{k} = \frac{(E_{jp}^{k})^{-1}}{(E_{jp}^{k})^{-1} + (E_{jn}^{k})^{-1}}$
\State $\alpha_{jn}^{k} = \frac{(E_{jn}^{k})^{-1}}{(E_{jp}^{k})^{-1} + (E_{jn}^{k})^{-1}}$

\State \textbf{$\bm{\mathcal{L}_{PNB}}$ ($\bm{x_{i}^k}$, $\bm{y_{i}^k}$):} $\# $ Positive Negative Balanced loss 

\State \quad return $ - [\sum_{i=1}^{N^k}\sum_{j = 1}^{C}(\mu * \alpha_{jp}^{k} * (\alpha_{jp}^{k} * y_i^k * log(f(x_i^k)) + (\alpha_{jn}^{k} * (1 - y_i^k) * log(1 - f(x_i^k))))]$

\end{algorithmic}
\end{algorithm}

\begin{algorithm}[htb]
\caption{FedBB: \textbf{Fed}erated learning with PN\textbf{B} loss and C\textbf{B}R}
\label{alg:algorithm3}
\textbf{Input}: Private datasets $D = \{D_k| k \in \mathcal{K}\}(\mathcal{K} = \{1, \dots, K\})$, global model $\theta_g$, local models $\theta = \{\theta_k|k\in \mathcal{K}\}$, $k$ is the index of clients, $C$ is the number of classes, $K$ is the number of clients, $x_{ki}$ is the input data and $y_{ik}$ is the label in $i^{th}$ batch of client $k$.
\begin{algorithmic}[1] %[1] enables line numbers
\State \textbf{LocalLearning($\bm{D_k, \theta_g}$):}

\State \quad $\theta_k = \theta_g$

\State \quad \textbf{for} \textit{each local epoch e from 1 to E} \textbf{do}
\State \quad \quad \textbf{for} \textit{each local batch i from 1 to S} \textbf{do}
\State \quad \quad \quad $\theta^i_k = \theta^{i-1}_k - \nabla\mathcal{L}_{PNB}(x_{ki}, y_{ki})$

\State \quad \textbf{return} $\theta_k$
\State
\State \textbf{for} \textit{each model aggregation round t = 0, 1, \dots, T-1} \textbf{do}

\State \quad \textbf{for} \textit{each client k = 0, 1, \dots, K-1} \textbf{do}

\State \quad \quad $\theta_k^t =$ LocalLearning($\mathcal{D}_k, \theta^{t-1}_g$)

\State \quad $\theta^t_g$ = $\sum_{k=1}^{K}\mathcal{W}_k * \theta_k^t$

\end{algorithmic}
\end{algorithm}

\section{Convergence Bound of FedBB}
\label{Proof}

As well known, the objective function of federated learning is as below.
\begin{equation}
    \min_{w} \{ F(w) = \sum_{k=1}^{N}p_kF_k(w) \}
\end{equation}

In FedBB, the local objective is as below. $\mathcal{W}_k$ is the weight from the CBR and the local objective function is PNB loss. $n_k$ is the number of data in $k^{th}$ client and $x_{k, j}$ is the $j^{th}$ input sample of $k^{th}$ client. $\omega$ in this equation is the weight parameter. 

\begin{equation}
    F_k(w) = \mathcal{W}_k \sum_{j=1}^{n_k} \mathcal{L}_{PNB}(w;x_{k, j})
\end{equation}

\cite{61} assumes that the stochastic gradient has a limit and this value impacts the overall convergence result bound. 

\begin{equation}
E||\triangledown F_k(w_t^k,\xi) - \triangledown F_k (w_t^k) ||^2 \leq \sigma_k^2
\end{equation}

This means that when we train the local model, the loss value has an upper bound. In the case in which data are highly skewed, it is well known that the average loss value of the minority class is higher than the majority class \cite{68}. 

\begin{equation}
    E[\triangledown F(\omega_t^k,\xi_s) \leq E[\triangledown 
 F(\omega_t^k,\xi_m)]
\end{equation}

$s$ indicates the minority class and $m$ indicates the majority class. Based on this, we can formulate an expansion of the Lipschitz condition.

\begin{equation}
    ||E[\triangledown F(\omega_t^k,\xi_s)] - E[\triangledown 
 F(\omega_t^k,\xi_m)]|| \leq M * ||p(\xi_s) - p(\xi_m)||
\end{equation}

$M$ is the scaling factor depending on several conditions such as \textcolor{black}{the} amount of data and domain gap inside a class. From \textcolor{black}{the} aforementioned conditions, we can see that $\sigma_k^2$ increases as the data \textcolor{black}{becomes} more skewed. However, when we \textcolor{black}{apply} PNB loss, we can decrease the upper bound by re-weighting which lowers the impact of majority classes and increases the impact of minority classes.

\begin{equation}
    ||E[\triangledown F(\omega_t^k,\xi_s)] - E[\triangledown 
 F(\omega_t^k,\xi_m)]|| \leq M * ||\alpha * p(\xi_s) - \beta * p(\xi_m)|| \leq M * ||p(\xi_s) - p(\xi_m)||
\end{equation}

$\alpha$ is the up-weighting and $\beta$ is the down-weighting terms. As a result, we can rewrite the assumption as below. \textit{The equal sign holds when the dataset is completely balanced.}

\begin{equation}
E||\triangledown F_k(w_t^k,\xi_t^k) - \triangledown F_k (w_t^k) ||^2 \leq \acute{\sigma}_k^{2} \leq \sigma_k^2
\end{equation}

Based on this assumption, we can also rewrite the Lemma 2 in \cite{61} as below.
\begin{equation}
\begin{split}
E|| g_t - \Bar{g_t} || & = E || \sum_{k=1}^N p_k(\triangledown F_k(w_t^k,\xi_t^k) - \triangledown F_k(w_t^k)) ||^2 \\
                       & = \sum_{k=1}^Np_{k}^{2} E|| \triangledown F_k (w_t^k, \xi_t^k) -\triangledown F_k (w_t^k) ||^2 \\
                       & \leq \sum_{k=1}^N p_k^2 \acute{\sigma}_k^2 \leq \sum_{k=1}^N p_k^2 \sigma_k^2
\end{split}
\end{equation}

Finally, we can rewrite the convergence result bound. $\triangle_t = E || \Bar{w}_t - w^*||^2$. The variables used in \textcolor{black}{the} below equation are \textcolor{black}{the} same as \cite{61}.

\begin{equation}
\begin{split}
\triangle_{t+1} \leq (1 - \eta_t\mu)\triangle_t + \eta_t^2\acute{B} \\
    (\Acute{B} = \sum_{k=1}^{N}p_k^2\acute{\sigma}_k^2 + 6L\Gamma + 8(E - 1)^2G^2)
\end{split}
\end{equation}

\begin{equation}
\begin{split}    
E[F(w_T)] - F^* & \leq \frac{\mathcal{K}}{\gamma + T - 1}(\frac{2\acute{B}}{\mu} + \frac{\mu\gamma}{2}E||w_1 - w^*||^2)\\
               & \leq \frac{\mathcal{K}}{\gamma + T - 1}(\frac{2B}{\mu} + \frac{\mu\gamma}{2}E||w_1 - w^*||^2)
\end{split}
\end{equation}

The theoretical underpinning of convergence bounds serves as a pivotal benchmark in evaluating the minimum guaranteed performance of an algorithm. Such bounds inherently delineate the convergence rate, establishing a direct correlation between the tightness of the bound and the efficiency of convergence. Specifically, in the context of achieving predefined performance objectives, an algorithm characterized by a more stringent convergence bound is theoretically predisposed to attain said objectives with greater alacrity compared to its counterparts possessing more lenient bounds. Consequently, the comparatively lower convergence bound associated with FedBB not only furnishes a theoretical substantiation of its accelerated convergence (which can conserve resources) compared to FedAvg but also suggests superior robustness, generalizability across diverse datasets, and enhanced efficiency. These attributes collectively underscore the significance of convergence bounds as a critical factor in the optimization and practical deployment of federated learning algorithms.

\section{Contribution of FedBB to Healthcare Community}
\textcolor{black}{It is widely known that FL occupies an important position in the healthcare domain because of its inherent privacy and data ownership issues. However, although a large amount of work is proposed to optimize FL, relatively little work is specialized in the medical domain.} Most of the FL methods are developed based on natural image datasets and multi-class classification in cross-device settings.

FL in the medical domain shares some common properties which are different from the others. First, FL in the medical domain is cross-silo FL. It has a relatively small number of clients who participate every round when there is no unexpected incident such as a blackout. In addition, because medical data is hard to \textcolor{black}{produce}, the data scale is relatively small compared to the natural image data. Because the probability of disease occurrence varies by disease, the global dataset, which combines all clients' datasets, shows a long-tailed distribution.

Furthermore, medical images have different characteristics compared to natural images \cite{62}. Medical images are grayscale images that have one channel or three or four channels with identical values for each channel. Intensity value plays a key role in medical image whereas local contrast is more important when leveraging natural image. Medical images are also more homogeneous than natural images because they share the same anatomy. 

Regarding the task, medical image classification is usually multi-label classification because one patient can have several diseases. 

Because of the aforementioned attributes, it is common that an algorithm that performs effectively on natural images in a cross-device setting fails to function well on medical images or in the cross-silo setting as shown in our experiments \cite{8, 11, 12, 27}.

FedBB, through extensive experiments, demonstrated its superiority on medical X-ray datasets \textcolor{black}{and} natural image datasets in the cross-silo setting. Therefore, we expect to be able to satisfy the demand for FL methodologies specific to the medical field.


\end{document}